\documentclass[conference]{IEEEtran}
\IEEEoverridecommandlockouts
\usepackage{cite}
\usepackage{amsmath,amssymb,amsfonts}
\usepackage{algorithmic}
\usepackage{graphicx}
\usepackage{textcomp}
\usepackage{xcolor}
\def\BibTeX{{\rm B\kern-.05em{\sc i\kern-.025em b}\kern-.08em
    T\kern-.1667em\lower.7ex\hbox{E}\kern-.125emX}}

\usepackage[acronym]{glossaries}
\usepackage{glossaries-extra}
\usepackage[super]{nth}
\setabbreviationstyle[acronym]{long-short}
\newacronym{ai}{AI}{Artificial Intelligence}
\newacronym{ml}{ML}{Machine Learning}
\newacronym{rl}{RL}{Reinforcement Learning}
\newacronym{dl}{DL}{Deep Learning}
\newacronym{ot}{OT}{Operational Technology}
\newacronym{cps}{CPS}{Cyber-Physical System}
\newacronym{plc}{PLC}{Programmable Logic Controller}
\newacronym{hil}{HiL}{Hardware-in-the-Loop}
\newacronym{ppo}{PPO}{Proximal Policy Optimization}
\newacronym{dqn}{DQN}{Deep Q-Learning}
\newacronym[plural=pn,firstplural=Petri Nets (PNs)]{pn}{PN}{Petri Net}
\newacronym[plural=ics,firstplural=Industrial Control Systems (ICS)]{ics}{ICS}{Industrial Control System}

\makeatletter 
\newcommand{\linebreakand}{%
  \end{@IEEEauthorhalign}
  \hfill\mbox{}\par
  \mbox{}\hfill\begin{@IEEEauthorhalign}
}
\makeatother 

\usepackage{tikz}
\tikzset{>=latex}
\usepackage{url}

\usepackage{cleveref}
\crefname{section}{Sect.}{Sect.}
\Crefname{section}{Section}{Sections}
\crefname{figure}{Fig.}{Fig.} 

\begin{document}

\title{Deep Q-Learning versus Proximal Policy Optimization: Performance Comparison in a Material Sorting Task\\
\thanks{Reuf Kozlica and Simon Hirländer are supported by the Lab for Intelligent Data Analytics Salzburg (IDA Lab) funded by Land Salzburg (WISS 2025) under project number 20102-F1901166-KZP.}
}

\author{

\IEEEauthorblockN{Reuf Kozlica}
\IEEEauthorblockA{\textit{Information Technologies and Digitalisation} \\
\textit{Salzburg University of Applied Sciences}\\
Salzburg, Austria \\
reuf.kozlica@fh-salzburg.ac.at}

\and
\IEEEauthorblockN{Stefan Wegenkittl}
\IEEEauthorblockA{\textit{Information Technologies and Digitalisation} \\
\textit{Salzburg University of Applied Sciences}\\
Salzburg, Austria \\
stefan.wegenkittl@fh-salzburg.ac.at}

\linebreakand 

\IEEEauthorblockN{Simon Hirländer}
\IEEEauthorblockA{\textit{Artificial Intelligence and Human Interfaces} \\
\textit{Paris Lodron University Salzburg}\\
Salzburg, Austria \\
simon.hirlaender@plus.ac.at}

}

\maketitle

\begin{abstract}
This paper presents a comparison between two well-known deep \gls{rl} algorithms: \gls{dqn} and \gls{ppo} in a simulated production system. We utilize a \gls{pn}-based simulation environment, which was previously proposed in related work. The performance of the two algorithms is compared based on several evaluation metrics, including average percentage of correctly assembled and sorted products, average episode length, and percentage of successful episodes. The results show that \gls{ppo} outperforms \gls{dqn} in terms of all evaluation metrics. The study highlights the advantages of policy-based algorithms in problems with high-dimensional state and action spaces. The study contributes to the field of deep \gls{rl} in context of production systems by providing insights into the effectiveness of different algorithms and their suitability for different tasks.
\end{abstract}


\section{Motivation}
There is a plethora of simulation environments used for \gls{rl} tasks. Among the most popular ones are MuJoCo, Gazebo, Webots, Gymnasium and PyBullets \cite{koerber2021simulationenvironments}. Most of the well-known environments rely on a physics engine, where the real world is being portrayed on a very low abstraction level. Designing such a simulation environment on this abstraction level can be a difficult and time consuming task. \glspl{pn} represent a mathematical modelling language used to describe and simulate systems and processes. This way the simulation is being designed on a fairly high abstraction level, which helps to reduce the design complexity of such a system. In this work, we use \glspl{pn} as a basis for solving \gls{rl} tasks as shown in \cite{petrinet2020hu,riedmann2022petrinetsimulation, harb2022strategies} and as been used in \cite{schaefer2023deploying_rl}.

The main bottleneck in programming this kind of production tasks is the effectiveness of algorithms being used, thus this paper examines two different modern approaches on solving the \gls{rl} problem: value approximation and policy approximation.

\section{Related Work}
\label{sec:related_work}
Riedmann et al. conducted an experiment applying traditional Q-learning to a task using a \gls{pn} model as a simulation environment. They showed that a \gls{pn} simulation model can be a suitable basis for training a \gls{rl} agent \cite{riedmann2022petrinetsimulation} and presented strategies for developing a supervisory controller using deep \gls{rl} in a production context \cite{harb2022strategies}.

They designed a case study including a simple sorting task on a material flow production line shown in \cref{fig:model_factory}. The facility consisted of an entry point, rotary table, assembly station, storage and exit. At the entry point the lower and upper part of the product which is to be transported are placed on a transport carriage. Afterwards, the product is transported to the assembly station using conveyor belts and the rotary table. At the assembly station a rivet is being installed, which connects the lower part of the product with the upper part. The product is labeled either blue or green. The green products are transported to the storage and blue products are to be transported to the exit station. 

They have shown that using traditional Q-learning, the agent has been able to correctly assembly and sort approximately 1.5 products of 3 given in each episode.

\section{State of Research}
This section provides a comprehensive review of the current literature on \gls{rl}, \gls{dqn} and \gls{ppo}.
\label{sec:state_of_the_art}
\subsection{Reinforcement Learning}
\gls{rl} is a type of machine learning that seeks to emulate the way humans learn, particularly in their early years when exploration is crucial. It involves learning through interaction with an environment, making it a computational approach to learning~\cite{sutton2018reinforcement}. 

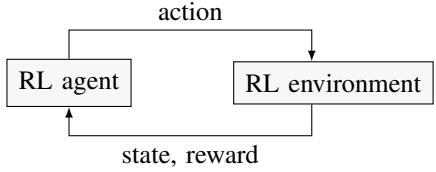
\begin{figure}[ht]
  \centering
  \begin{tikzpicture}[%
      boxnode/.style = {draw, inner sep=1ex, fill=black!3},
      node distance=10em]
    \node[boxnode]                (agent) {RL agent};
    \node[boxnode,right of=agent] (env)   {RL environment};

    \draw[->] (agent) -- ++(0,2em)  -| node[pos=0.25,above] {action}        (env.135);
    \draw[<-] (agent) -- ++(0,-2em) -| node[pos=0.25,below] {state, reward} (env.225);
  \end{tikzpicture}

  \caption{Standard Reinforcement Learning Setup.}
  \label{fig:rl_setup_basic}
\end{figure}

In a typical \gls{rl} setup (as depicted in \cref{fig:rl_setup_basic}), there are two main components: the agent and the environment. The agent is in a particular state $s \in S$ and can perform actions $a \in A$, which may belong to either discrete or continuous sets and can be multi-dimensional~\cite{kober2013reinforcement}. The action performed by the agent changes the state of the environment, causing a transition and is evaluated by the environment, which sends a reward $R_t$ to the agent after each step. The agent's objective is to maximize the accumulated reward over time. During the interaction with the environment, the agent tries to find a policy $\pi$ that maps states to actions, and maximizes the overall return $G = \sum_{k=0}^{\infty}\gamma^kR_{k+1}$, where $\gamma < 1$~\cite{kaelbling1996survey}. The policy can be either deterministic or probabilistic, with the former using the same action for a given state $a = \pi(s)$, while the latter maps a distribution over actions when in a specific state~$a \sim \pi(s,a)$.

\subsection{Deep Q-Learning (\gls{dqn})}
The idea of Q-learning has been introduced in 1989 by Watkins and proven in detail by \cite{qlearning1992watkins}. It is based on a so called state-action value function, also known as Q-function. This function of a given policy $\pi$, $Q^{\pi}(s,a)$, represents the expected return of a trajectory starting from a specific state $s$, taking a specific action $a$ and following the policy $\pi$ thereafter. The optimal policy $Q^{*}(s,a)$ is defined as the maximum return that can be achieved, when starting from state $s$, taking an action $a$ and following the optimal policy thereafter. Using the Bellman optimality equation performing an iterative update of the Q-values, the optimal Q-function can be approximated. It has been shown that this way the $Q_i \rightarrow Q^*$ as $i \rightarrow \infty$ \cite{dqn2013mnih}. 

In order to perform the iterative update of the Q-function, a Q-table, containing all pairs of $s$ and $a$, must be used. Dependent on the application the state space and action space can become large, causing out-of-core situations, thus being impractical for usage. Instead of using a Q-table, Deep Q-learning trains a neural network to estimate the Q-values, which acts as a function approximator. This way the neural network with parameters $\theta$ serves as an approximate of the optimal function: $Q(s,a;\theta) \approx Q^*(s,a)$.

Using nonlinear function approximators in the \gls{rl} domain is known to be unstable and can even diverge when such an approximator is used to represent the Q-function \cite{functionapproximation1997tsitsiklis}. To tackle this problem, a biologically inspired mechanism named experience replay has been introduced. The agent's experiences $e_t = (s_t,a_t,r_t,s_{t+1})$ are stored at each time-step $t$ in a data set $D_t = \{e_1, ..., e_t\}$. In the training process, the Q-learning updates are performed using mini-batches of experience drawn uniformly randomly from the pool of samples named replay buffer \cite{dqn2013mnih}.

Using Deep Q-learning, authors of \cite{Mnih2015HumanlevelCT} evaluated the performance of the same network for 49 different tasks on the Atari 2600 platform, which is designed to be difficult and engaging for human players. This agent outperformed other existing \gls{rl} methods on 43 games and reached the level of professional human game testers on all 49 games.

\subsection{Proximal Policy Optimization (\gls{ppo})}
Another approach for solving \gls{rl} problems using neural networks as function approximators have been policy gradient methods. \gls{ppo} is trying to tackle the known issues which other methods like vanilla policy gradient method (VPG) and trust region policy optimization (TRPO)\cite{schulman2015trpo} are dealing with: Vanilla policy gradient methods struggle with data efficiency and robustness and TRPO is fairly computationally expensive, as well as not suited for architectures that include noise \cite{schulman2017ppo}.

The second-order optimization used by TRPO makes it computationally complex and difficult to scale up for large scale problems \cite{wang2019trulyppo}. 
\gls{ppo} is trying to improve those methods by introducing an algorithm which retains the data efficiency and performance of TRPO but only using first-order optimization.

The main contribution of this novel approach is the clipped surrogate objective, loss function which is to be optimized:

\begin{equation}
L^{CLIP}(\theta) = \hat{\mathbb{E}}_t[min(r_t(\theta)\hat{A}_t, clip(r_t(\theta), 1-\epsilon, 1+\epsilon)\hat{A}_t)]
\end{equation}

In the above equation, an expectation of minimum of two terms is being computed. The first term inside the minimum operator $r_t(\theta)\hat{A}_t$ is the standard policy gradient objective. $r_t(\theta)$ is defined as probability ration between the action under the current policy and the same action under the old policy $\frac{\pi_{\theta}(a_t|s_t)}{\pi_{\theta_{old}}(a_t|s_t)}$. The second term clips the probability ratio, removing the possibility of moving $r_t$ outside the interval $[1-\epsilon, 1+\epsilon]$, where $\epsilon$ is a hyperparameter. In case of a positive advantage $\hat{A}_t > 0$, meaning that selected action has a positive effect on the outcome, the loss function flattens out when $r_t(\theta)$ gets too large. The clipped objective prevents taking a too far step from the current policy. Moreover, when the advantage is negative $\hat{A}_t < 0$ the loss function flattens out when $r_t(\theta)$ converges to zero, meaning particular action is less likely on current policy.

\begin{figure*}[t]
    \centerline{\includegraphics[width=1.0\linewidth]{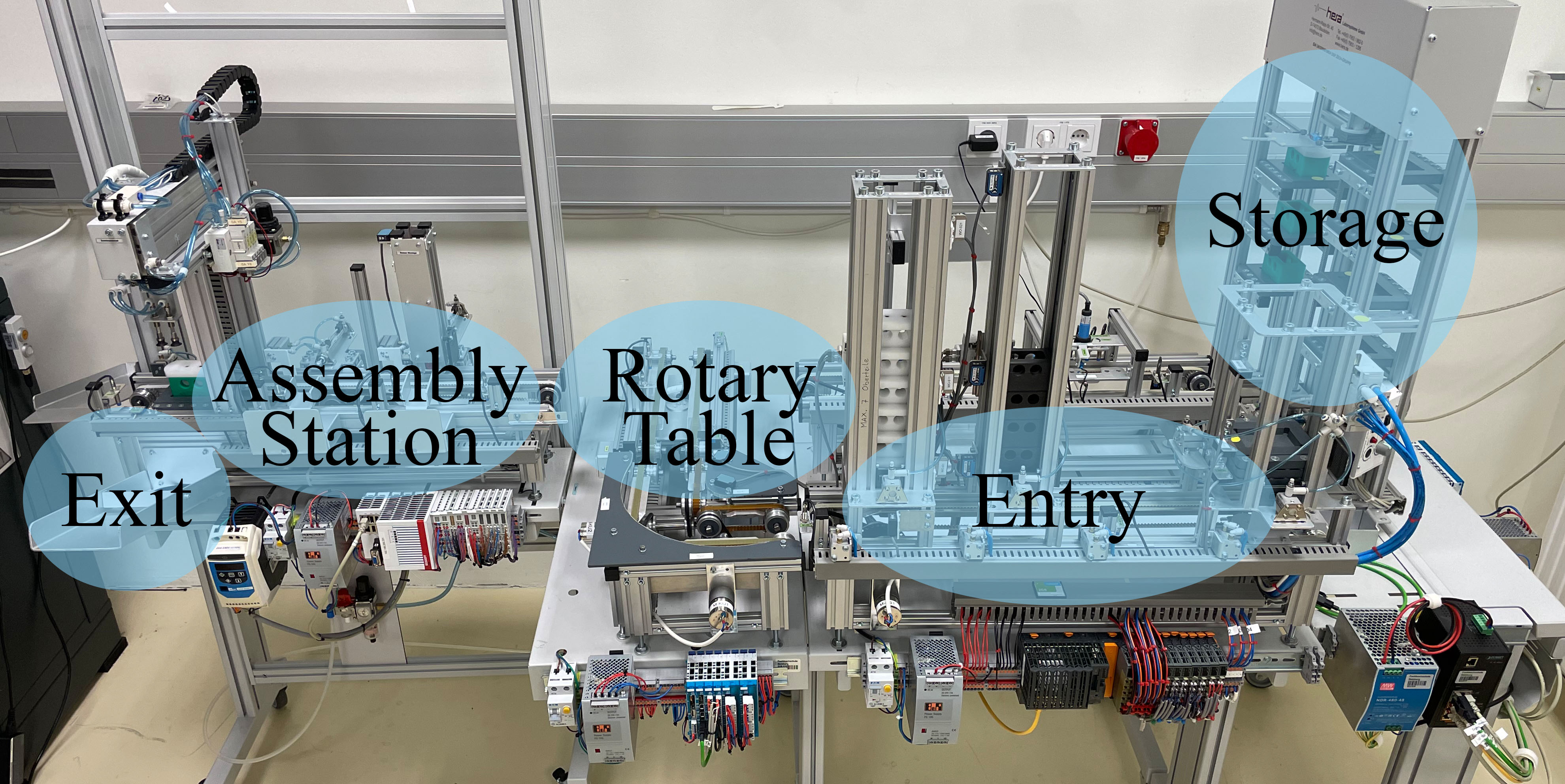}}
    \caption[Model factory placed in the Smart Factory Lab]{Model factory placed in the Smart Factory Lab\footnotemark.}
    \label{fig:model_factory}
\end{figure*}

\subsection{Petri Nets (\glspl{pn})}
\glspl{pn} constitute a well-known paradigm for specification and operation of different systems, which is able to process multiple independent activities at the same time. This ability differentiates \glspl{pn} from finite state machines, which are only able to operate one single state at a given moment. Petri nets thus are widely used as a tool for modelling, simulation, analysis and control of automated manufacturing systems \cite{petri2019seatzu}. 
Following the mathematical definition of PNs, a developed set of methodology and approaches is available in literature for theoretical and practical analysis of PNs~\cite{petri2021grobelna}.

A \gls{pn} is a bipartite graph whose vertices consist of places (represented by circles) and transitions (represented by bars). Places and transitions are directly connected by arcs while tokens (usually represented by black dots) describe its state. Furthermore, two places are not allowed to be connected nor are two transitions allowed to be directly connected. A \gls{pn} can formally be denoted as a quadruple $N = (P, T, Pre, Post)$, where: $P = {p_1, ..., p_m}$ represents a finite set of places, $T={t_1, ..., t_n}$ is the finite set of transitions, and $Pre, Post: P \times T \rightarrow \mathbb{N}^{m \times n}$ are the pre- and post-incidence matrices that define direct arcs from places to transitions and from transitions to places, respectively \cite{petri2019seatzu}. Since 1962, when they have been proposed by Carl Adam Petri, \glspl{pn} have evolved to meet needs of different systems, some of them include timed-, coloured-, interpreted-, stochastic-, and fuzzy models.

\section{Use Case and Simulation Environment}
\label{sec:usecase_simulation_environment}
For this work the same simulation environment based on \glspl{pn} proposed by \cite{riedmann2022petrinetsimulation} as described in \cref{sec:related_work} has been used. The original contribution of this paper is the comparison of two well-known algorithms in the world of \gls{rl}. Deep Q-learning has been chosen as a representative of value based deep \gls{rl} algorithms and \gls{ppo} as a well-known representative of policy based deep \gls{rl} algorithms. 

The case study, as proposed by \cite{riedmann2022petrinetsimulation}, has been created using a blue print of a model factory shown in \cref{fig:model_factory}. As introduced in \cref{sec:related_work}, this factory consists of 5 modules, namely: entry, rotary table, assembly station, storage and exit. The entry storage stores three different types of parts as illustrated in \cref{fig:assembled_product}. The first part is the transport carriage, which is used for moving goods on the conveyor belts. The transport carriage is moving the other two parts which are stored in the entry storage: lower and upper part of the product. Moreover, the entry is also responsible for the assembly of the listed parts. After the parts have been assembled at the entry point, the goods are to be transported to the rotary table where they can be transported either to the storage point, to the assembly station or to the exit. 

The model factory is designed to be representative for a real production system. Hence, it includes aspects like the transportation of goods using conveyor belts and a rotary table. Also the product modification is represented through the assembly station. Lastly, the storage of goods is found in both entry point and the main storage unit.

\subsection{Task Definition}\label{subsec:task_definition}
\begin{figure}[htb]
    \centerline{\includegraphics[width=1.0\linewidth]{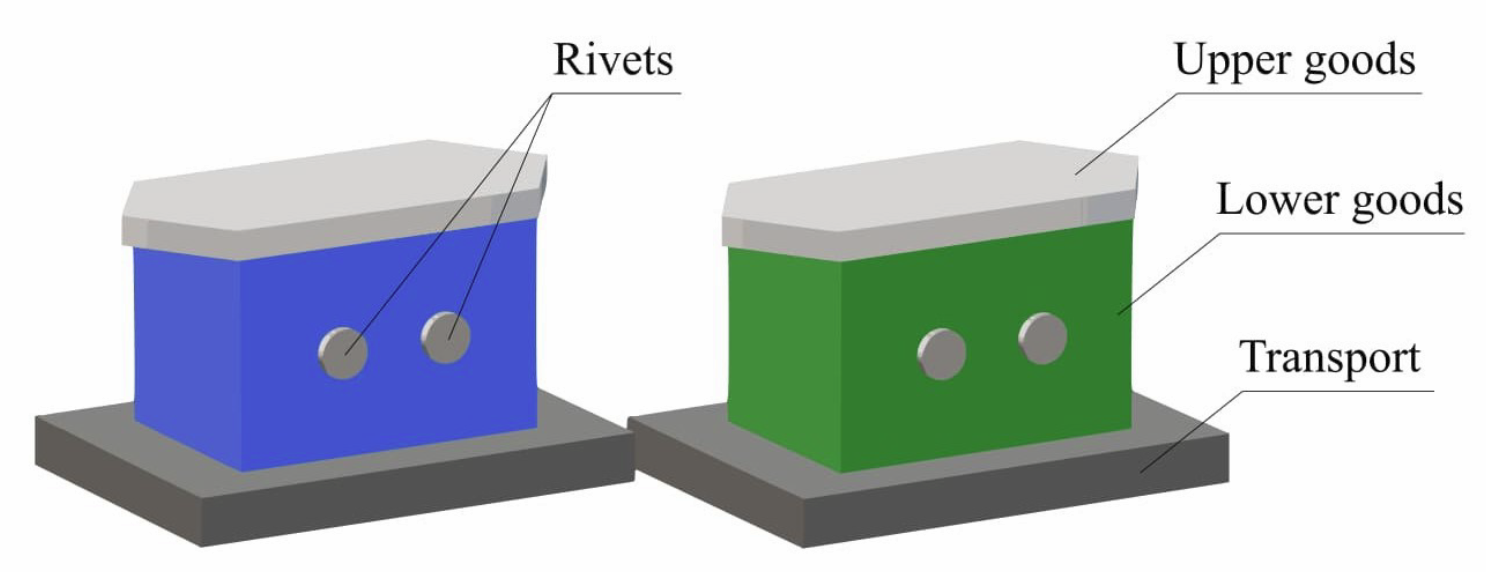}}
    \caption{Assembled products placed on the transport carriage.}
    \label{fig:assembled_product}
\end{figure}

The task formulated in this use case is to transport and assemble the goods through this model factory following defined rules. Therefore, a specific number of goods $G=\{g_1, g_2, ..., g_{n_{max}}\}$ of a random color are being placed at the entry point during each simulation run. First, the goods are to be transported to the assembly station. At the assembly station two rivets are being installed into each product, as depicted in \cref{fig:assembled_product}, connecting the upper part to the lower part of the product. After the rivets have been installed, the blue products must be transported to the exit, while the green products are to be transported to the storage. 

\footnotetext{\url{https://its.fh-salzburg.ac.at/forschung/smart-factory-lab/}}

\subsection{Simulation}
The \gls{pn} based simulation proposed by \cite{riedmann2022petrinetsimulation} and extended in \cite{harb2022strategies} has been used for the realization of this task. This model has been implemented in Python using the SNAKES toolbox \cite{pommereau2015snakes} which has been slightly extended to facilitate the timed transitions. Thus, each transition is assigned a hidden place $p_{h}$ enabling modelling the relative difference in execution times. 

\section{Implementation}
This section details the implementation of the suggested case study introduced in \cref{sec:usecase_simulation_environment}. 

\subsection{State and Action Space}
The action space of the agent is defined by the \gls{pn} transitions to be controlled. Similar to \cite{zinn2021hrl_robotic_devices}, a dedicated action is being assigned to every \gls{pn} transition $T$ turning it on or off. Since there may be states where no transition should be fired, for example when firing a transition would result in a collision later, a non-action has been added. Respectively, the action space consists of 12 actions encoded as a vector $\Vec{a}=[a_1, a_2, ..., a_{12}]^{T}$, $a_i \in \{0, 1\}$. The number $\#A$ of actions is thus depending on the number of actuators in the given production line.

The state space comprises all places $P$ contained in the \gls{pn} model. In order to translate the \gls{pn} state to the state space of the agent, it is necessary to consider the similarities between places. Therefore, all storage places are encoded using a scalar value, since different tokens in storage places do not require separate treatment based on the color of the product. However, all other places are one-hot encoded, as tokens must be handled differently based on their color. One-hot vectors are generated with a number of components corresponding to the number of possible distinct token values for the respective place. Thus, the resource places number of components in the encoding is 2, meaning resource available or unavailable. Moreover, the number of components of all other places is either 4 or 6, depending on whether assembled tokens can reach these places. Additionally, the hidden places $p_h$ are also one-hot encoded with 5 as a number of components and included within the state space. Hence, the state has been encoded as a vector $\vec{s}=[s_1, s_2, ..., s_{101}]^T$, $s_i \in \mathbb{N}_0$ consisting of 101 components, as shown in \cref{tab:state_space_encoding}.

\begin{table}[htbp]
\caption{State Space Encoding}
    \begin{center}
        \begin{tabular}{c|c|c}
        \textbf{Parameter} & \textbf{Count}& \textbf{Number of components} \\
        \hline
        \hline
        Resource place & 2 & 2 \\
        \hline
        Storage place & 4 & 1 \\
        \hline
        Regular place & 5 & 6 \\
        \hline
        Regular short place & 2 & 4 \\
        \hline
        Hidden place & 11 & 5 \\
        \hline
        \end{tabular}
        \label{tab:state_space_encoding}
    \end{center}
\end{table}

\subsection{Reward Design}
\label{subsec:reward_design}

The goal of the \gls{rl} agent is to autonomously learn to transport and assemble the goods according to the task formulation described in \cref{subsec:task_definition}. For the purpose of this paper two slightly different reward functions $R_1$ and $R_2$ are defined and compared. Reward
$$
R_1 = \begin{cases}
    -1 & \text{collision occurred}\\
    -0.5 & \text{product incorrectly sorted}\\
    -0.01 & \text{invalid transition fired}\\
    +0 & \text{(non-)transition fired}\\
    +1 & \text{goal reached}
\end{cases}
$$
only focuses on the task of correctly assembling and sorting the given products. Therefore, a simple transition is not being penalized $r_{transition} = 0$. A detected collision in the production system represents an event which shall not occur under any circumstances and results in a negative reward of $r_{collision} = -1$. Products which are incorrectly sorted lead to a negative reward of $r_{incorrect} = -0.5$. Furthermore, transitions which do not affect the production system, i.e. firing a transition of an unoccupied place, are penalized with a small negative reward $r_{invalid} = -0.01$. The successful completition of the task leads to a reward of $r_{successful} = 1$.

The reward function $R_2$ however is also considering the number of transitions needed for the completition of the task, hence defining the reward $r_{transition} = -0.001$. It is important to keep this negative reward sufficiently small, since a disproportional increase would result in the agent constantly choosing the non-action, which gets no reward, all the time. The reward values are chosen iteratively by comprehensive experiments.
$$
R_2 = \begin{cases}
    -1 & \text{collision occurred}\\
    -0.5 & \text{product incorrectly sorted}\\
    -0.01 & \text{invalid transition fired}\\
    -0.001 & \text{(non-)transition fired}\\
    +1 & \text{goal reached}
\end{cases}
$$

\begin{figure*}[ht]
    \centerline{\includegraphics[width=1.0\linewidth]{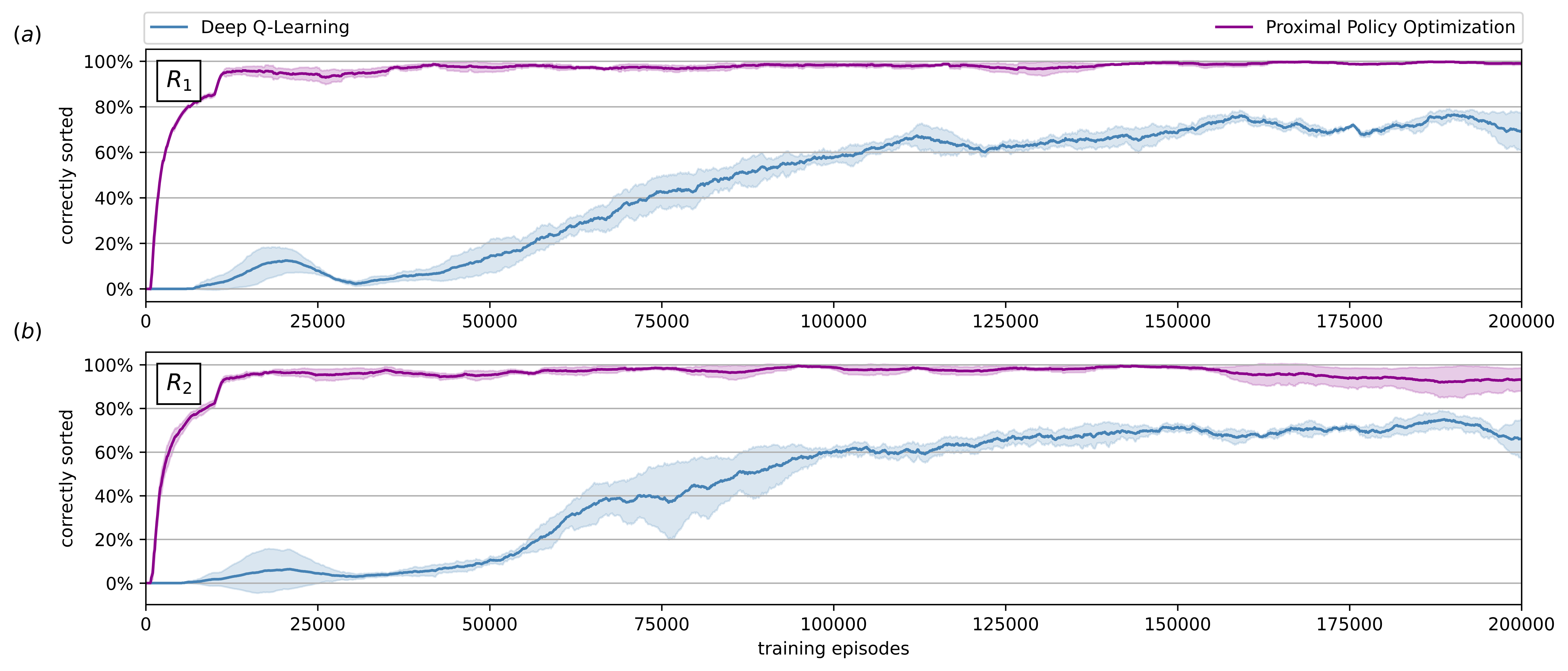}}
    \caption{Training process of Deep Q-Learning and Proximal Policy Optimization for rewards $R_1$ and $R_2$. After every 100 training episodes, the agent is evaluating the policy for 5 episodes. The evaluation results are smoothed for 200 data points.}
    \label{fig:training}
\end{figure*}

\section{Evaluation}
To show the applicability of the described modelling method to a wider range of \gls{rl} algorithms, the model is implemented and tested using \gls{dqn} and \gls{ppo} algorithms described in~\cref{sec:state_of_the_art}. Both algorithms are specified to use a similar structure of the neural network. The dimensions of the approximation network are defined as following: input layer consisting of the state vector $\vec{s}$ with 101 components, two hidden layers with 200 nodes in the first layer and 100 nodes in the second layer and an output layer given by $\vec{a}$, consisting of 12 actions. Through an iterative process, the number of nodes within the neural network's layers was systematically reduced until the most efficient configuration capable of solving the given task was discovered, thereby optimizing the network's computational efficiency and minimizing its complexity.

The $\epsilon$-greedy strategy, chosen as an exploration strategy for \gls{dqn}, quantifies the probability that the agent will select a random action instead of choosing the action with the highest predicted value according to the current estimate of the Q-function (i.e., acting greedily). The exploration rate is initialized to 1 at the onset of training, and subsequently reduced linearly, in order to balance the agent's exploration and exploitation tendencies as it learns to navigate the environment. This progressive reduction in ensures that the agent transitions from a more exploratory behavior to a more exploitative one as its policy becomes more refined and robust, ultimately culminating in a minimum exploration rate of $0.1$.

To ensure the comparability of the results both \gls{ppo} and \gls{dqn} are trained for 200,000 episodes having a maximum of 100 time steps for each episode. On an off-the-shelf hardware setup the full training process takes approximately 24 hours for the \gls{dqn} and 40 hours for \gls{ppo}. Each setup has been trained with 5 different seeds and all performance plots are showing the mean of all runs as a solid line, while the semi-transparent filled area shows the standard deviation.

\subsection{Analysis of Reward Design}
The analysis of reward design is a critical aspect of \gls{rl} research. It refers to the process of selecting and designing the appropriate reward function that an algorithm uses to evaluate its performance and learn from it. The right reward function can significantly impact the performance and effectiveness of a reinforcement learning algorithm. 

\cref{fig:training} shows the training process of both algorithms \gls{dqn} and \gls{ppo} for both rewards $R_1$ and $R_2$, described in \cref{subsec:reward_design}. The training process plot clearly shows that the \gls{ppo} algorithm outperformed the \gls{dqn} algorithm in terms of learning speed for both reward functions. This indicates that \gls{ppo} was able to learn and adapt to the task more efficiently and effectively than \gls{dqn}.

\begin{figure}[ht]
    \centerline{\includegraphics[width=1.0\linewidth]{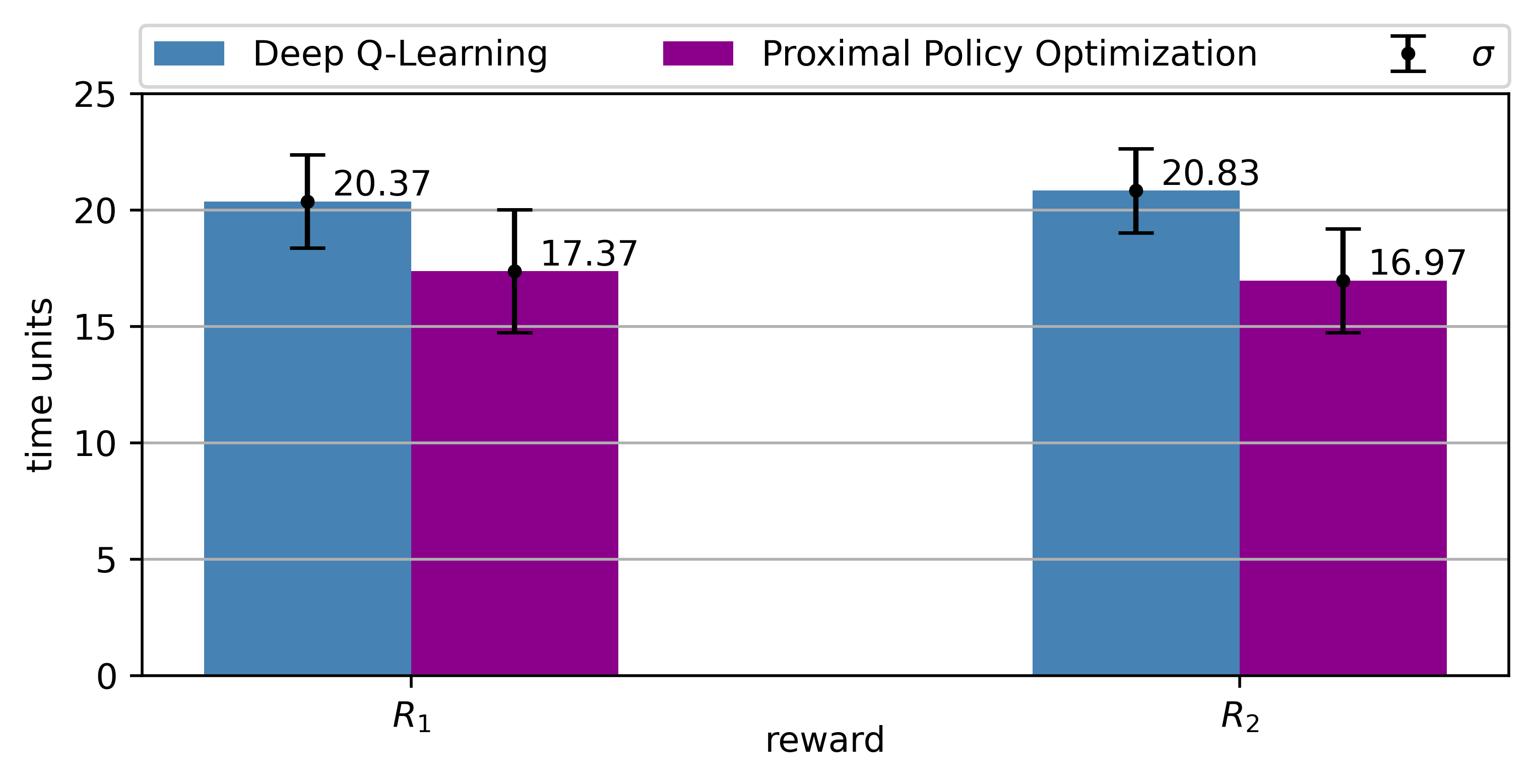}}
    \caption{Comparison of the learned policy using \gls{dqn} and \gls{ppo}: Time steps needed to complete the sorting task successfully.}
    \label{fig:time_evaluation}
\end{figure}

Nevertheless, a direct comparison of the training performance of \gls{ppo} in \cref{fig:training} $(a)$ and $(b)$ shows that even a small change in the reward design can cause instabilities of the learned policy. Especially when looking into the last 40,000 episodes in $(b)$, where the policy is getting slightly worse. This could be caused by a phenomenon known as catastrophic forgetting, which can be overcome by not updating the model if the performance is getting worse.

Finally, all 5 trained policies for both algorithms in combination with $R_1$ and $R_2$ are evaluated for 100 episodes with a random sequence of products to be assembled and sorted. \cref{fig:time_evaluation} shows the time the agent needs to correctly complete the task. For the \gls{ppo} algorithm, the design of the $R_2$ reward function achieved to reduce the time needed to successfully accomplish the task. Specifically, the time for \gls{ppo} was 0.4 time units faster. On the other hand, for the \gls{dqn} algorithm, the same design did not have the same effect.

\begin{figure}[ht]
    \centerline{\includegraphics[width=1.0\linewidth]{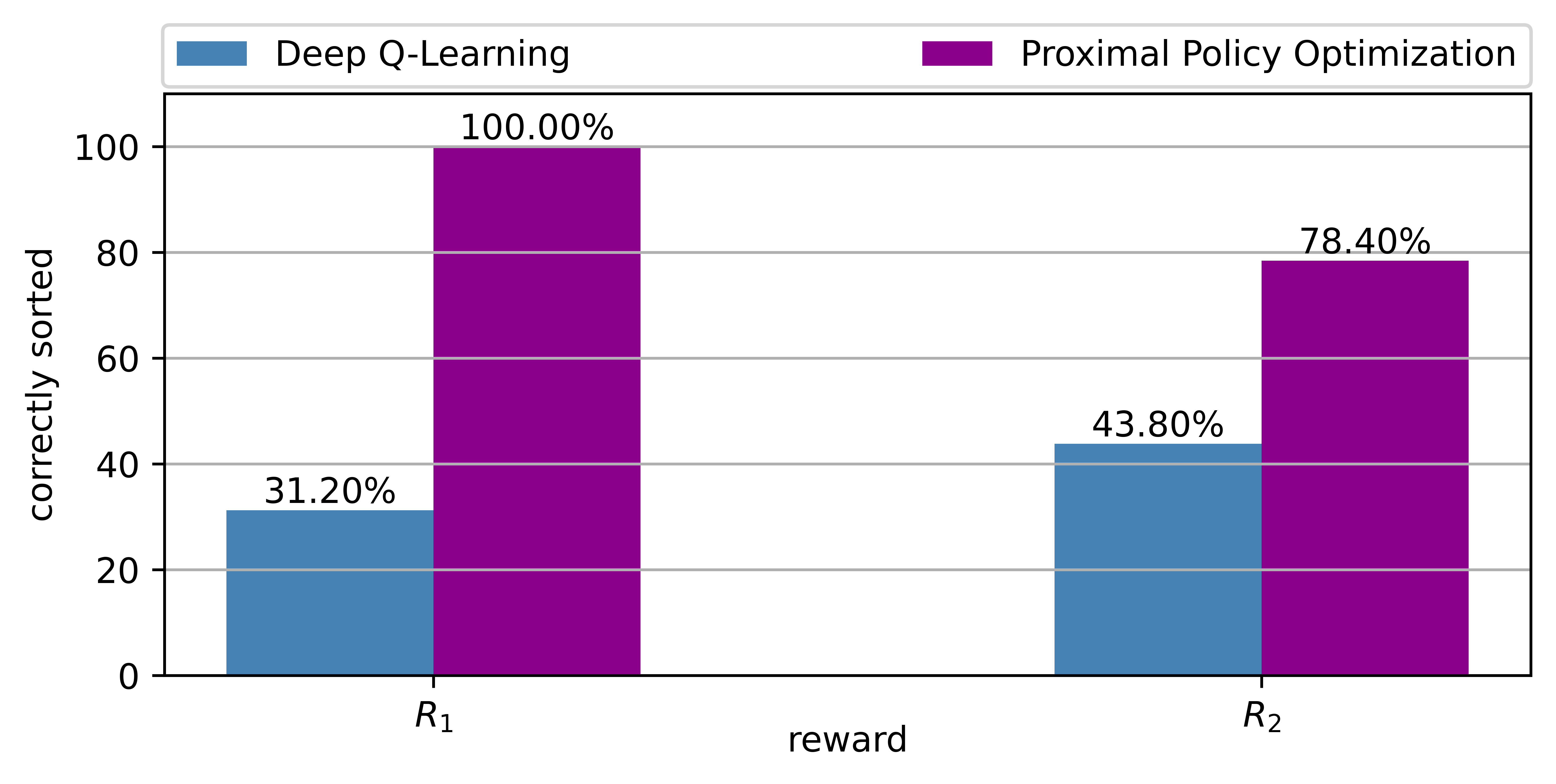}}
    \caption{Comparison of the policy trained using \gls{dqn} and \gls{ppo}: Percentage of episodes demonstrating successful sorting and assembling of all products by the \gls{rl} agent for $R_1$ and $R_2$ reward designs.}
    \label{fig:correctly_sorted}
\end{figure}

The evaluation of trained policies in terms of the percentage of correctly sorted and assembled products during an episode is presented in \cref{fig:correctly_sorted}. The results demonstrate a notable performance advantage of the \gls{ppo} algorithm compared to the \gls{dqn} algorithm for both reward designs.

However, further analysis of the results showed that the $R_2$ reward design actually managed to improve the performance of the \gls{dqn} algorithm, while it had the opposite effect on \gls{ppo}. These findings suggest that the impact of reward design on the performance of reinforcement learning algorithms is not always straightforward and can vary depending on the specific algorithm and task at hand.

\section{Conclusion and Future Work}
The results highlight the importance of choosing the right algorithm for the task and the benefits of using more advanced reinforcement learning methods like \gls{ppo}, removing the need of manual reward shaping presented by \cite{harb2022strategies}. 

Nevertheless, the model has limitations that must be considered. The model assumes a fixed number of products in each simulation run, which may not reflect the real-world scenario accurately. Additionally, the model is limited to a maximum of 100 time steps, which may not capture the long-term effects of the manufacturing process.
Moving forward, our research will focus on overcoming the aforementioned limitations to enhance the model's applicability in real-world settings. We will also explore the impacts of factory scaling by comparing conventional \gls{rl} approaches with hierarchical methods.

\bibliographystyle{IEEEtran}
\bibliography{references}

\end{document}